\documentclass[10pt,twocolumn,letterpaper]{article}

\usepackage{multirow}%
\usepackage[pagenumbers]{cvpr} 
\usepackage{float}

\definecolor{cvprblue}{rgb}{0.21,0.49,0.74}
\usepackage[pagebackref,breaklinks,colorlinks,allcolors=cvprblue]{hyperref}

\def\paperID{} 
\def\confName{CVPR}
\def\confYear{2026}

\title{NeuroSeg Meets DINOv3: Transferring 2D Self-Supervised Visual Priors to 3D Neuron Segmentation via DINOv3 Initialization}

\author{
Yik San Cheng \quad Runkai Zhao$^{*}$ \quad Weidong Cai$^{*}$ \\
The University of Sydney \\
{\tt\small yiksan.cheng@sydney.edu.au \quad runkai.zhao@sydney.edu.au \quad tom.cai@sydney.edu.au}
} 

\begin{document}
\maketitle
\renewcommand{\thefootnote}{}
\footnotetext{* Corresponding authors.}
\renewcommand{\thefootnote}{\arabic{footnote}}


\begin{abstract}
2D visual foundation models, such as DINOv3, a self-supervised model trained on large-scale natural images, have demonstrated strong zero-shot generalization, capturing both rich global context and fine-grained structural cues. However, an analogous 3D foundation model for downstream volumetric neuroimaging remains lacking, largely due to the challenges of 3D image acquisition and the scarcity of high-quality annotations. To address this gap, we propose to adapt the 2D visual representations learned by DINOv3 to a 3D biomedical segmentation model, enabling more data-efficient and morphologically faithful neuronal reconstruction. Specifically, we design an inflation-based adaptation strategy that inflates 2D filters into 3D operators, preserving semantic priors from DINOv3 while adapting to 3D neuronal volume patches. In addition, we introduce a topology-aware skeleton loss to explicitly enforce structural fidelity of graph-based neuronal arbor reconstruction. Extensive experiments on four neuronal imaging datasets, including two from BigNeuron and two public datasets, NeuroFly and CWMBS, demonstrate consistent improvements in reconstruction accuracy over SoTA methods, with average gains of 2.9\% in Entire Structure Average, 2.8\% in Different Structure Average, and 3.8\% in Percentage of Different Structure. Code: \href{https://github.com/yy0007/NeurINO}{https://github.com/yy0007/NeurINO}.
\end{abstract}

\begin{figure}
  \centering
  \centerline{\includegraphics[width=8.1cm]{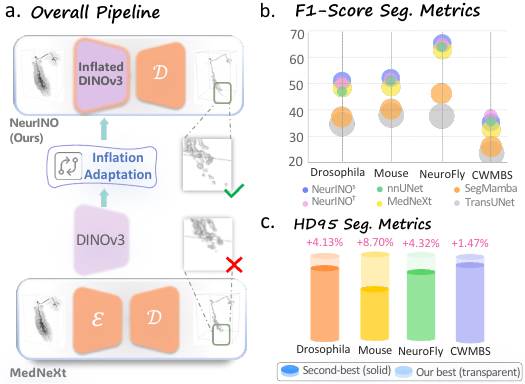}}
  \vspace{-6pt}
\caption{
  \textbf{a.} We propose a data-efficient 3D neuron segmentation model, namely NeurINO, bridging a 2D foundation model (DINOv3) with 3D volumetric neuroimaging;
  \textbf{b.} {F1-score comparison} across four neuronal datasets, where bubble size indicates model parameters. NeurINO achieves consistent segmentation improvements with comparable model complexity;  
  \textbf{c.} NeurINO marginally outperforms the second-best methods (MedNeXt and nnUNet) by 1–9\%, achieving the lowest HD95 across all datasets.
  }
\label{fig:Teaser}
\vspace{-20pt}
\end{figure}

\vspace{-20pt}
\section{Introduction}
\label{sec:intro}

Accurate reconstruction of neuronal morphology from mesoscale volumetric light microscopy is essential for mapping brain connectivity and function, requiring tracing long-projecting, thin axonal and dendritic neurites across brain regions to preserve circuit topology ~\cite{manubens2023bigneuron,gao2023single,liu2016rivulet,liu2024single,peng2021morphological,qiu2024whole,chen2020computer,wang2021ai,tavakoli2025light,liu2025connectivity}. Yet the extreme slenderness, tortuosity, deep branching hierarchies, and dense interweaving of neurites, compounded by low SNR and anisotropic resolution, make it difficult for volumetric segmentation and tracing to preserve both topological completeness and morphological continuity. 
Early automated methods~\cite{yang2019fmst,chen2015smarttracing,radojevic2019automated,xiao2013app2,liu2018automated,peng2011automatic,tang2017automatic,zhang2016reconstruction} relied on handcrafted algorithms and heuristics, reducing manual workload but suffering from brittle priors and limited cross-dataset or -species generalization. With the advent of deep learning, recent approaches~\cite{wang2019multiscale,li20193d,san2025dineuro, zhao2023pointneuron} improve adaptability by treating segmentation as a data-driven pre-processing step that extracts neuron masks from low-SNR microscopy volumes. Nevertheless, despite progress catalyzed by large-scale community efforts such as DIADEM~\cite{brown2011diadem}, BigNeuron~\cite{peng2015bigneuron,peng2015diadem}, and databases such as NeuroMorph.Org~\cite{tecuatl2024accelerating} and NeuroXiv~\cite{jiang2025neuroxiv}, the performance is limited by the scarcity of high-quality annotations. 

While recent label-efficient paradigms, including semi-supervised learning, self-supervised pre-training, and foundation model transfer \cite{xie2024pairaug, wang2024rethinking, butoi2023universeg, wu2024one, huang2023rethinking, xu2024hybrid, wang2024enhancing, Yi2025DualFete, gao2025dino, li2025meddinov3, liu2025does, zhao2025segment, an2025raptor}, have reduced annotation costs in biomedical tasks, they remain insufficient for neuron reconstruction. The critical limitation lies in a misalignment of objectives: these methods typically optimize for voxel-level accuracy and lack explicit structural supervision for the complex, branching topology and thin, intertwined geometry of neuron arbors. Therefore, existing models may yield high segmentation metrics but fail to preserve topological fidelity, producing erroneously fragmented or missed structures.    

With big data emergence, large language models (LLMs) and Vision Foundation Models (VFMs) exhibit surprising zero-shot generalization and serve as foundational priors for downstream tasks. In 2D vision, foundation models broadly span three fold: (i) image-only self-supervision, (ii) vision–language alignment, and (iii) task-oriented pretraining. Web-scale data learning yields transferable embeddings, while lightweight adaptation (e.g., LoRA, prompt tuning) further reduces label dependence and improves out-of-domain robustness. In biomedical imaging, however, progress remains largely 2D-centric \cite{huang2021gloria, bannur2023learning, wu2023medklip, zhang2023biomedclip, li2023llavamed}. Image–text models trained on multimodal corpora \cite{zhang2023biomedclip, li2023llavamed, yin2026afire}, together with 2D self-supervised and promptable segmentation backbones \cite{ma2024segment, archit2025segment,stevens2024bioclip, ryu2025vision, wang2025interpretable, ding2025multimodal}, perform well to slice-based recognition, retrieval, and segmentation across organs and tissues. By contrast, volumetric and microscopy-scale modalities remain underexplored due to costly acquisition, restricted data access, and labor-intensive voxel/skeleton-level annotation, especially for fine neuronal morphology. Despite growing interest in community, a generalizable 3D biomedical foundation model has yet to emerge \cite{zhuang2025bio2vol}.

To address data scarcity and the lack of foundational priors in volumetric neuroimaging, we re-think to decomposing the volumetric learning problem into 2D intra-slice feature extraction and 3D inter-slice aggregation. Concretely, we first initialize the 3D neuron segmentation encoder from "inflating" the pre-trained weights from a large-scale 2D foundation model (e.g., DINOv3~\cite{simeoni2025dinov3}). This strategy immediately endows the model with robust, web-scale priors for intra-slice semantics (edges, textures, spatial patterns) before it encounters any 3D data. To bridge the intricate dimensional gap, we design an inflation-based adaptation strategy that lifts these 2D weights to 3D version, allowing the model focus exclusively on learning inter-slice correlations, such as structural continuity, topology, and long-range projections, rather than redundantly relearning basic visual features. To further guide cross-slice spatial integration and enforce morphological fidelity, we propose a novel topology-aware skeleton loss, explicitly penalizing structural discontinuities and topological errors. Consequently, our method reduces the sample learning complexity and enables data-efficient morphological learning boosted by robust 2D foundational knowledge (Figure~\ref{fig:Teaser}). The major contributions of this work are summarized as follows: 

\begin{itemize}    

    \item We propose a data-efficient 3D \underline{\textbf{Neur}}on segmentation model initialized from D\underline{\textbf{INO}}v3-learned visual priors, namely as \underline{\textbf{NeurINO}}, effectively bridging large-scale 2D foundation models with 3D volumetric neuroimaging.
   
    \item We design a simple but efficient inflation-based adaptation strategy to lift 2D derived weights to 3D version, transferring robust priors for intra-slice semantics in microscopic volume. Also, we introduce a topology-aware skeleton loss to learn on inter-slice correlations, explicitly penalizing morphological discontinuities and errors.

    \item Extensive experiments conducted on BigNeuron, NeuroFly and CWMBS, demonstrate that our method outperforms SoTA methods with average gains of 2.9\% in ESA, 2.8\% in DSA, and 3.8\% in PDS.

\end{itemize} 
\begin{figure*}

  \centering
  \centerline{\includegraphics[width=\linewidth]{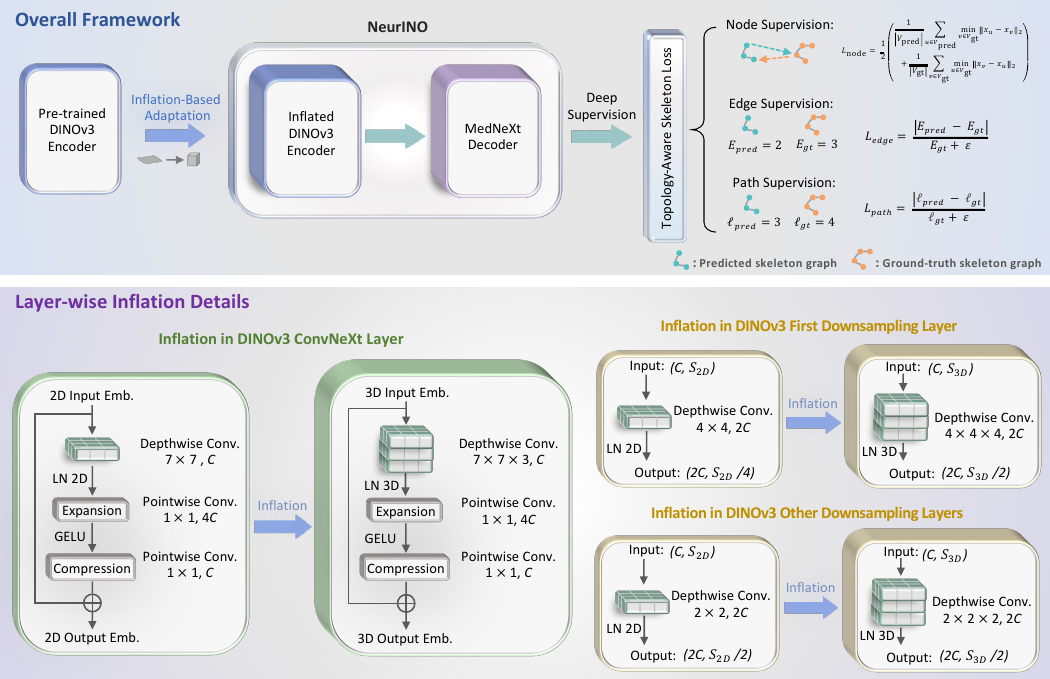}}
    \vspace{-6pt}
   \caption{\textbf{Overview of our framework.} We adapt a pretrained DINOv3 ConvNeXt encoder for volumetric neuron segmentation through an inflation-based 3D adaptation strategy, which spatially expands 2D convolutional kernels into 3D while preserving pretrained semantics. The inflated encoder is coupled with a symmetric MedNeXt-style decoder to recover fine-grained neuronal morphology. Multi-level outputs are supervised jointly with the proposed Topology-Aware Skeleton Loss (TASL), which measures structural discrepancies between predicted and ground-truth skeleton graphs. $S_{2D}$ and $S_{3D}$ denote the spatial sizes in 2D and 3D feature maps, respectively.}

   \label{fig:network}
   \vspace{-15pt}
\end{figure*}

\section{Related Work}
\label{sec:formatting}

\subsection{Neuronal Morphological Learning}
Reconstructing 3D neuronal morphology from volumetric microscopy has long been a core challenge in neuroscience. Traditional methods relied on handcrafted heuristics, including seed-point dependent tracing~\cite{yang2019fmst, chen2015smarttracing, radojevic2019automated} and seed-independent algorithms~\cite{xiao2013app2, liu2018automated, peng2011automatic, tang2017automatic, zhang2016reconstruction}. Despite their utility, these methods suffer from rigid priors and limited adaptability.  
With the advent of deep learning, 3D encoder-decoder networks (e.g., 3D U-Net~\cite{cciccek20163d}, V-Net~\cite{milletari2016v}) have markedly improved neuron segmentation through data-driven feature learning. Subsequent refinements~\cite{wang2019multiscale, wang2021voxel, cheng2026modeling, zhao2023pointneuron} further enhanced spatial representation and semantic consistency. However, these approaches remain constrained by limited data availability and lack explicit topological supervision, leading to disconnected or morphologically inconsistent neuron reconstructions.

\subsection{Topology-Aware Loss}
Accurate segmentation of tubular or network-like structures requires preserving topology beyond voxel-level accuracy. Prior works have pursued this through specialized architectures~\cite{cheng2021joint, lin2023dtu, mosinska2019joint}, persistent-homology supervision~\cite{clough2020topological, wong2021persistent}, and differentiable skeletonization~\cite{shit2021cldice, menten2023skeletonization}. Lightweight variants, such as Skeleton Recall Loss~\cite{kirchhoff2024skeleton}, enhance structural continuity by penalizing missed centerlines. However, most of these approaches focus on skeleton overlap or centerline fidelity without modeling graph-level structural attributes of branching architectures. In contrast, our topology-aware skeleton loss explicitly enforces node, edge, and path-level consistency, providing fine-grained supervision over morphological continuity and completeness.




\section{Method}

We posit that data-efficient 3D neuron segmentation is best achieved by decomposing the problem of 2D intra-slice feature representation and 3D inter-slice aggregation. The intra-slice component, tasked with identifying low-level features (e.g., edges, textures), can leverage existing 2D foundation models (e.g., DINOv3), as re-learning these from scarce 3D data is redundant and inefficient.
Our pipeline, as shown in Figure~\ref{fig:network}, operationalizes this by first initializing the 3D encoder via "inflating" pre-trained 2D weights. This injects robust, web-scale visual priors before any 3D training occurs. We introduce a bespoke inflation-based adaptation strategy, detailed in Section~\ref{Section3.1}, a critical step that directs the model’s learning capacity toward modeling inter-slice structural continuity. However, standard voxel-wise objectives are topology-agnostic and thus insufficient for filament-like geometry of neurons; they cannot effectively penalize fragmentation or erroneous mergers. To impose this essential higher-order structural constraint, we introduce the Topology-Aware Skeleton Loss (TASL), detailed in Section~\ref{Section3.2}. TASL functions as an explicit topological regularizer, ensuring the learned 3D aggregation produces morphologically faithful reconstructions.

\subsection{Inflation-Based Adaptation} 
\label{Section3.1}

To bridge the 2D-3D disparity, we adopt an inflation-based adaptation strategy inspired by~\cite{cheng2024boosting}, which spatially inflates the 2D convolution kernels from DINOv3 into their 3D counterparts. This design enables full 3D spatial reasoning for volumetric neuron segmentation while preserving the pretrained semantic priors from DINOv3. Formally, each 2D kernel with weights 
\(W_{2D} \in \mathbb{R}^{C_{out} \times C_{in} \times k_h \times k_w}\) 
is converted into a 3D kernel 
\(W_{3D} \in \mathbb{R}^{C_{out} \times C_{in} \times k_d \times k_h \times k_w}\)
where \(C_{in}\) and \(C_{out}\) denote the input and output channel dimensions, and 
\(k_h\), \(k_w\), and \(k_d\) correspond to the kernel height, width, and depth along the axial dimension, respectively.
We implement two complementary inflation schemes: center inflation and average inflation.
In the center inflation scheme, the original 2D kernel is placed at the central slice of the inflated 3D kernel volume:



\begin{equation}
W_{3D}[:,:,c,:,:] = W_{2D}, \quad c = \left\lfloor \frac{k_d}{2} \right\rfloor ,
\end{equation} 
while all other slices are set to zero, i.e., $W_{3D}[:,:,t,:,:] = 0$ for $t \neq c$. This initialization preserves the spatial semantics of pretrained filters by embedding the 2D kernel at the center of the 3D volume.  
In average inflation, the 2D kernel is uniformly replicated along the depth dimension:


\begin{equation}
W_{3D}[:,:,t,:,:] = \frac{1}{k_d} W_{2D}, \quad t = 0, \dots, k_d-1,
\end{equation}

\noindent smoothing depth-wise receptive fields while preserving the statistical characteristics of the original filters. Both inflation schemes retain DINOv3’s visual knowledge and enable volumetric feature learning. The inflated encoder thereby inherits semantic richness and high-fidelity local representations from large-scale pretraining. 





\begin{table*}[t]
\centering
\scriptsize
\setlength{\tabcolsep}{10pt}
\renewcommand{\arraystretch}{1}
\caption{
\textbf{Quantitative comparison of neuron segmentation performance.}
Each cell reports {F1-score (\%) / HD95} (lower is better for HD95) across four datasets.
Columns correspond to datasets and rows to different segmentation methods.
{NeurINO$^{T}$} and {NeurINO$^{S}$} denote the Tiny and Small variants of NeurINO.
Best results are highlighted in \textcolor{red!70!black}{\textbf{red}}, and the second best in \textcolor{blue!70!black}{\textbf{blue}}.
}
\label{tab:seg_only_f1_hd95}

\setlength{\heavyrulewidth}{0.06em}
\setlength{\lightrulewidth}{0.05em}
\begin{tabular}{lcccccc}
\toprule
\textbf{Method} & \textbf{Source} & \textbf{Params (M)} & \textbf{Drosophila} & \textbf{Mouse} & \textbf{NeuroFly} & \textbf{CWMBS} \\
\midrule
nnUNet~\cite{isensee2021nnu}      & \textit{Nat. Methods 2021} & 27.66 & 47.20 / 3.20 & 52.05 / 10.12 & 63.36 / 18.33 & {{36.50 / 16.34}} \\
MedNeXt~\cite{roy2023mednext}     & \textit{MICCAI 2023} & 61.97 & {47.74 / 3.15} & 50.61 / 13.77 & 62.50 / 19.23 & 33.46 / 18.37 \\

\midrule
NeurINO$^{T}$ & - & 39.21 & \textcolor{blue!70!black}{\textbf{50.06 / 3.07}} & \textcolor{blue!70!black}{\textbf{52.50 / 9.50}} & \textcolor{blue!70!black}{\textbf{65.23}} / \textcolor{red!70!black}{\textbf{16.38}} & \textcolor{red!70!black}{\textbf{36.77 / 16.10}} \\

NeurINO$^{S}$ & - & 61.52 & \textcolor{red!70!black}{\textbf{50.19 / 3.02}} & \textcolor{red!70!black}{\textbf{52.73 / 9.24}} & \textcolor{red!70!black}{\textbf{65.44}} / \textcolor{blue!70!black}{\textbf{16.53}} & \textcolor{blue!70!black}{\textbf{36.55 / 16.27}} \\
\bottomrule
\end{tabular}
\end{table*}


\begin{table*}[t]
\centering
\small
\setlength{\tabcolsep}{4.5pt}
\renewcommand{\arraystretch}{2.5}
\caption{
\textbf{Quantitative comparison of neuron tracing performance.}
Each cell reports {ESA / DSA / PDS} (lower is better for all metrics) for two tracing algorithms: 
SmartTracing and NeuTube, evaluated across four datasets. 
{NeurINO$^{T}$} and {NeurINO$^{S}$} denote the Tiny and Small variants of NeurINO.
Best results are highlighted in \textcolor{red!70!black}{\textbf{red}}, and the second best in \textcolor{blue!70!black}{\textbf{blue}}. 
}
\label{tab:tracing_results}
\resizebox{\textwidth}{!}{
\fontsize{15}{10}\selectfont
\begin{tabular}{lcccccccc}
\toprule
\multirow{2}{*}{\textbf{Method}} &
\multicolumn{2}{c}{\textbf{Drosophila}} &
\multicolumn{2}{c}{\textbf{Mouse}} &
\multicolumn{2}{c}{\textbf{NeuroFly}} &
\multicolumn{2}{c}{\textbf{CWMBS}} \\
\cmidrule(lr){2-3} \cmidrule(lr){4-5} \cmidrule(lr){6-7} \cmidrule(lr){8-9}
& \textit{SmartTracing} & \textit{NeuTube}
& \textit{SmartTracing} & \textit{NeuTube}
& \textit{SmartTracing} & \textit{NeuTube}
& \textit{SmartTracing} & \textit{NeuTube} \\
\midrule

nnUNet~\cite{isensee2021nnu}
& 1.67 / 4.48 / \textcolor{red!70!black}{\textbf{0.20}} 
& 1.87 / 4.67 / 0.24
& 2.99 / 8.06 / \textcolor{blue!70!black}{\textbf{0.22}} 
& 5.36 / 16.10 / \textcolor{blue!70!black}{\textbf{0.22}}
& \textcolor{blue!70!black}{\textbf{22.78}} / 31.76 / 0.41 
& \textcolor{blue!70!black}{\textbf{4.22}} / 14.13 / \textcolor{red!70!black}{\textbf{0.15}}
& 36.93 / 42.10 / \textcolor{red!70!black}{\textbf{0.53}} 
& 8.02 / 14.73 / \textcolor{red!70!black}{\textbf{0.41}} \\

MedNeXt~\cite{roy2023mednext}
& 1.68 / 4.44 / \textcolor{red!70!black}{\textbf{0.20}} 
& 1.90 / 4.61 / 0.23
& 4.06 / 11.66 / 0.25 
& 6.81 / 19.61 / \textcolor{blue!70!black}{\textbf{0.22}}
& 27.52 / 34.74 / 0.40 
& 4.32 / 14.06 / 0.17
& 34.74 / 38.94 / 0.56 
& 7.93 / \textcolor{blue!70!black}{\textbf{13.58}} / \textcolor{blue!70!black}{\textbf{0.44}} \\



\midrule
NeurINO$^{T}$
& \textcolor{red!70!black}{\textbf{1.62}} / \textcolor{red!70!black}{\textbf{4.29}} / \textcolor{red!70!black}{\textbf{0.20}} 
& \textcolor{red!70!black}{\textbf{1.75}} / \textcolor{red!70!black}{\textbf{4.18}} / \textcolor{red!70!black}{\textbf{0.21}}
& \textcolor{red!70!black}{\textbf{2.81}} / \textcolor{blue!70!black}{\textbf{7.88}} / \textcolor{blue!70!black}{\textbf{0.22}} 
& \textcolor{blue!70!black}{\textbf{4.86}} / \textcolor{red!70!black}{\textbf{15.30}} / \textcolor{blue!70!black}{\textbf{0.22}}
& \textcolor{red!70!black}{\textbf{21.93}} / \textcolor{red!70!black}{\textbf{28.98}} / \textcolor{red!70!black}{\textbf{0.32}} 
& \textcolor{red!70!black}{\textbf{4.13}} / \textcolor{red!70!black}{\textbf{13.91}} / \textcolor{blue!70!black}{\textbf{0.16}}
& \textcolor{blue!70!black}{\textbf{34.09}} / \textcolor{red!70!black}{\textbf{37.76}} / \textcolor{blue!70!black}{\textbf{0.54}} 
& \textcolor{red!70!black}{\textbf{7.74}} / \textcolor{red!70!black}{\textbf{13.40}} / \textcolor{red!70!black}{\textbf{0.41}} \\

NeurINO$^{S}$
& \textcolor{blue!70!black}{\textbf{1.65}} / \textcolor{blue!70!black}{\textbf{4.40}} / \textcolor{blue!70!black}{\textbf{0.21}} 
& \textcolor{blue!70!black}{\textbf{1.84}} / \textcolor{blue!70!black}{\textbf{4.31}} / \textcolor{blue!70!black}{\textbf{0.22}}
& \textcolor{blue!70!black}{\textbf{2.92}} / \textcolor{red!70!black}{\textbf{7.73}} / \textcolor{red!70!black}{\textbf{0.21}} 
& \textcolor{red!70!black}{\textbf{4.73}} / \textcolor{blue!70!black}{\textbf{15.53}} / \textcolor{red!70!black}{\textbf{0.21}}
& 23.61 / \textcolor{blue!70!black}{\textbf{30.84}} / \textcolor{blue!70!black}{\textbf{0.36}} 
& 4.24 / \textcolor{blue!70!black}{\textbf{14.02}} / \textcolor{red!70!black}{\textbf{0.15}}
& \textcolor{red!70!black}{\textbf{33.48}} / \textcolor{blue!70!black}{\textbf{38.09}} / \textcolor{blue!70!black}{\textbf{0.54}} 
& \textcolor{blue!70!black}{\textbf{7.89}} / 13.67 / \textcolor{red!70!black}{\textbf{0.41}} \\
\bottomrule
\end{tabular}
}
\vspace{-15pt}
\end{table*}


\subsection{Topology-Aware Skeleton Loss}
\label{Section3.2}

Segmenting network-like structures such as neurons demands more than voxel-wise accuracy. 
While standard losses (e.g., Dice or cross-entropy) optimize local overlap, they fail to capture global connectivity and branching topology that determine morphological correctness. 
To mitigate this limitation, we introduce a Topology-Aware Skeleton Loss (TASL) as a regularization term that measures discrepancies between predicted and ground-truth skeleton graphs, while serving as an auxiliary signal for model selection. 

\noindent
\textbf{Skeleton-to-Graph Conversion.}
Given a predicted map 
$ p \in [0,1]^{B\times 2\times H_1\times H_2\times H_3}$ 
and ground truth 
$y\in\{0,1\}^{B\times 1\times H_1\times H_2\times H_3}$, 
we first obtain binarized segmentations:
\begin{equation}
\hat{y} = \mathbb{I}(p > \tau),
\end{equation}
where $\tau$ is a fixed threshold (e.g., $0.5$). 
We then extract morphological skeletons for the foreground using a skeletonization operator $\mathcal{S}(\cdot)$:
\begin{equation}
S_{\text{gt}} = \mathcal{S}(y), \qquad 
S_{\text{pred}} = \mathcal{S}(\hat{y}).
\end{equation}

Each skeleton mask is converted into a graph 
$G=(V,E)$ by treating each skeleton voxel as a node and connecting spatially adjacent voxels:

\begin{equation}
\begin{aligned}
V &= \{\mathbf{z}\in\mathbb{Z}^3 \mid S(\mathbf{z})=1\},\\
E &= \{(\mathbf{z}_i,\mathbf{z}_j)\mid \|\mathbf{z}_i-\mathbf{z}_j\|_2 \le r\},
\end{aligned}
\end{equation}

\noindent
where \(V\) and \(E\) denote the node and edge sets of the skeleton graph, respectively. Two skeleton voxels are connected if their Euclidean distance is within \(r = 2\) voxel units, which defines the local adjacency radius for graph construction.   
Let $(G_{\text{gt}}, G_{\text{pred}})$ denote the ground-truth and predicted graphs, 
with node coordinates 
$X_{\text{gt}}\!\in\!\mathbb{R}^{|V_{\text{gt}}|\times 3}$ and 
$X_{\text{pred}}\!\in\!\mathbb{R}^{|V_{\text{pred}}|\times 3}$.

\noindent
\textbf{Topological Discrepancies.}
TASL decomposes the topology mismatch into three complementary terms:

\noindent (1) Node-level discrepancy.
We compute a symmetric nearest-neighbor distance between node sets to quantify geometric deviation of branches:

\begin{equation}
\begin{aligned}
L_{\text{node}}(G_{\text{gt}}, G_{\text{pred}}) &=
\tfrac{1}{2}\!\Bigg(
\frac{1}{|V_{\text{pred}}|}\!\sum_{u\in V_{\text{pred}}}\min_{v\in V_{\text{gt}}}\!\|\mathbf{x}_u-\mathbf{x}_v\|_2 \\
&\quad + 
\frac{1}{|V_{\text{gt}}|}\!\sum_{v\in V_{\text{gt}}}\min_{u\in V_{\text{pred}}}\!\|\mathbf{x}_v-\mathbf{x}_u\|_2
\Bigg),
\end{aligned}
\end{equation}
which penalizes misplaced bifurcations and missing endpoints.

\noindent (2) Edge-level discrepancy. To measure global connectivity consistency, 
we compare the number of edges in the predicted and ground-truth graphs:

\begin{equation}
L_{\text{edge}}(G_{\text{gt}}, G_{\text{pred}}) 
= \frac{\big|\,|E_{\text{pred}}| - |E_{\text{gt}}|\,\big|}{|E_{\text{gt}}| + \varepsilon},
\end{equation}

\noindent
where $E_{\text{pred}}$ and $E_{\text{gt}}$ are the edge sets of the predicted and ground-truth graphs, respectively, and $\varepsilon$ is a small constant for numerical stability. This captures over- and under-connection errors at the graph level.

\noindent (3) Path-level discrepancy. To reflect long-range connectivity and branching completeness, we compare the mean connected-component lengths:

\begin{equation}
\mathcal{C}(G) = \{C_1, C_2, \dots, C_m\}, \quad
\ell(G) = \frac{1}{m}\sum_{i=1}^{m} |C_i|,
\end{equation}

\noindent
where $\mathcal{C}(G)$ denotes the set of connected components of $G$ and $|C_i|$ is the number of nodes in component $C_i$.
The discrepancy is then defined as:

\begin{equation}
L_{\text{path}}(G_{\text{gt}}, G_{\text{pred}}) =
\frac{\big|\ell(G_{\text{pred}}) - \ell(G_{\text{gt}})\big|}{\ell(G_{\text{gt}}) + \varepsilon},
\end{equation}

\noindent emphasizing structural continuity across long paths and penalizes disconnections or fragmentation.

\vspace{0.5em}
\noindent
\textbf{Overall Topological Loss.} 
The final topology-aware loss is defined as:
\begin{equation}
\mathcal{L}_{\text{TASL}} = 
\lambda_{\text{node}} L_{\text{node}} +
\lambda_{\text{edge}} L_{\text{edge}} +
\lambda_{\text{path}} L_{\text{path}},
\end{equation}
where $\lambda_{\text{node}}, \lambda_{\text{edge}}, \lambda_{\text{path}}$ 
control the relative importance of each topological component.

\subsection{Deep Supervision and Loss Functions}

Following standard practice in medical segmentation~\cite{isensee2021nnu}, we employ deep supervision to improve gradient flow and refine multi-scale features. 
Each output branch is supervised by a combination of Dice loss, cross-entropy (CE) loss, and our Topology-Aware Skeleton Loss (TASL). Total loss function is formulated as:







\begin{equation}
\mathcal{L}_{\text{total}} 
= \sum_{s=0}^{S} \lambda_s 
\big( 1 + \beta \, \mathcal{L}_{\text{TASL}}^s \big) 
\big( \mathcal{L}_{\text{Dice}}^s + \mathcal{L}_{\text{CE}}^s \big),
\end{equation}

\noindent
where $S$ denotes the number of deep supervision outputs, 
$\lambda_s$ is the weight assigned to each scale. The $\mathcal{L}_{\text{Dice}}^s$, $\mathcal{L}_{\text{CE}}^s$, and $\mathcal{L}_{\text{TASL}}^s$ 
represent the Dice loss, cross-entropy loss, and TASL computed at scale $s$, respectively. 
$\beta$ controls the relative strength of the TASL to the overall objective.





\begin{figure*}
  \centering
  \centerline{\includegraphics[width=16.5cm]{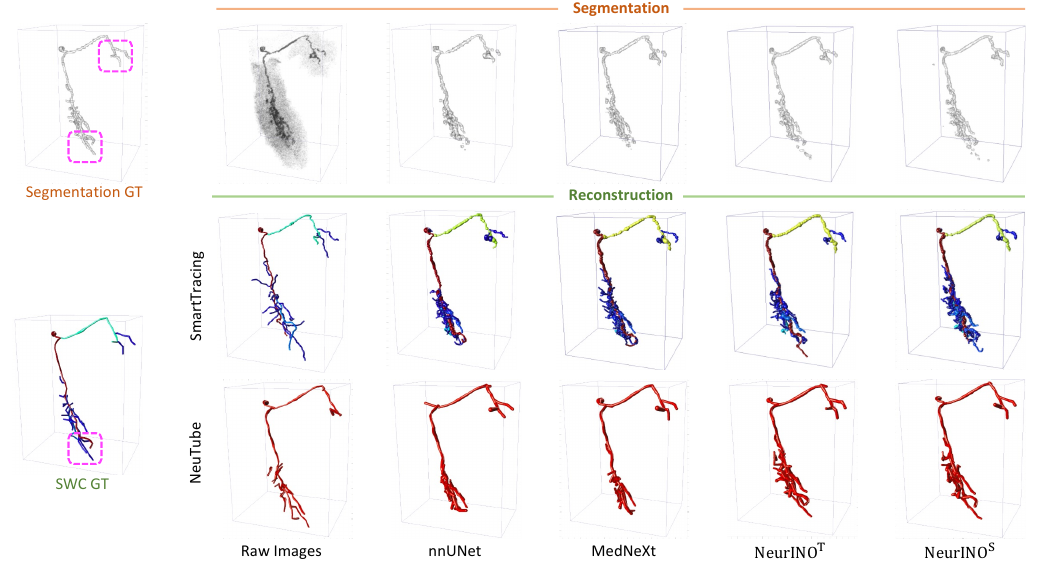}}
    \vspace{-6pt}
   \caption{\textbf{Visualization comparison of the segmentation and tracing results on the Drosophila dataset.} The top row presents raw images and segmentation outputs, followed by two rows showing the corresponding reconstruction results generated by SmartTracing and NeuTube. \textcolor{magenta}{\textbf{Magenta boxes}} indicate severe false negatives (missed neurites) in other methods. Best viewed in zoom-in regions.} 
   \label{fig:visual_Fly}
   \vspace{-4pt}
\end{figure*}

\begin{figure*}
  \centering
  \centerline{\includegraphics[width=17cm]{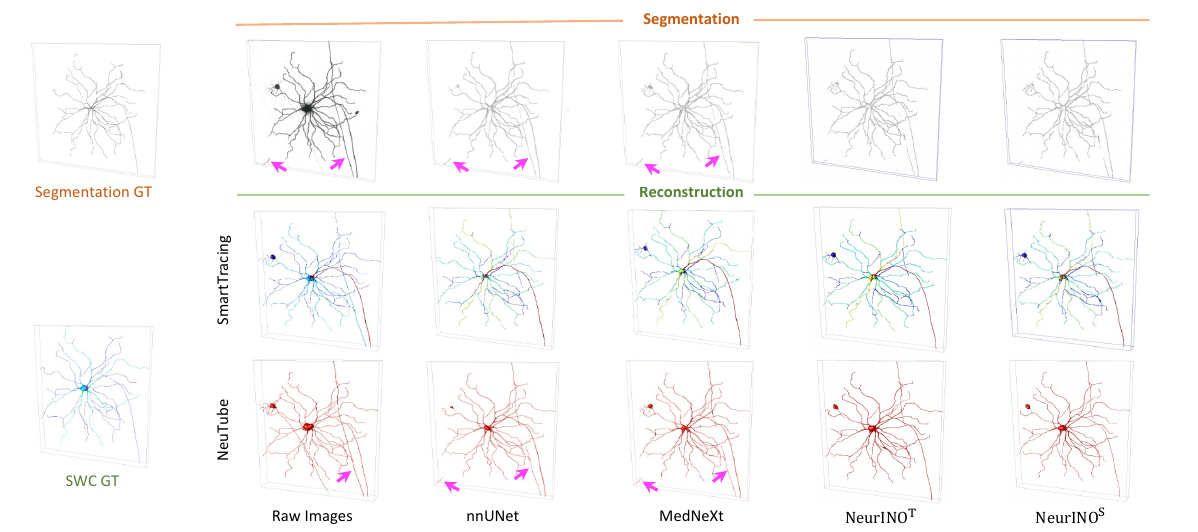}}
    \vspace{-6pt}
   \caption{\textbf{Visualization comparison of the segmentation and tracing results on the Mouse dataset.} The top row presents raw images and segmentation outputs, followed by two rows showing the corresponding reconstruction results generated by SmartTracing and NeuTube. \textcolor{Magenta}{\textbf{Magenta arrows}} highlight severe {false positives} in other methods. Best viewed in zoom-in regions.}
   \label{fig:visual_Mouse}
   \vspace{-15pt}
\end{figure*}

\begin{figure*}
  \centering
  \centerline{\includegraphics[width=17cm]{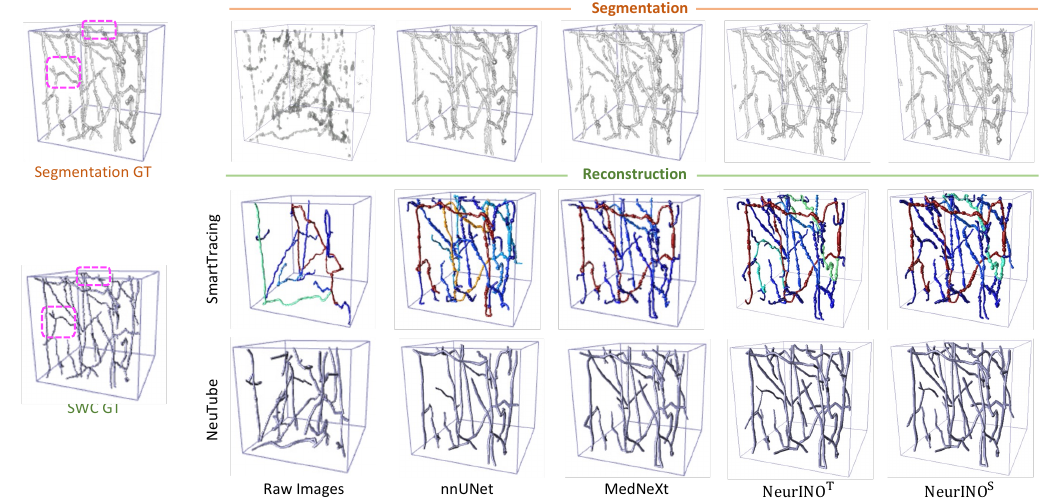}}
    \vspace{-6pt}
   \caption{\textbf{Visualization comparison of the segmentation and tracing results on the NeuroFly dataset.} The top row presents raw images and segmentation outputs, followed by two rows showing the corresponding reconstruction results generated by SmartTracing and NeuTube. \textcolor{magenta}{\textbf{Magenta boxes}} indicate severe false negatives (missed neurites) in other methods. Best viewed in zoom-in regions.} 
   \label{fig:visual_NeuroFly}
   \vspace{-4pt}
\end{figure*}

\begin{figure*}
  \centering
  \centerline{\includegraphics[width=17cm]{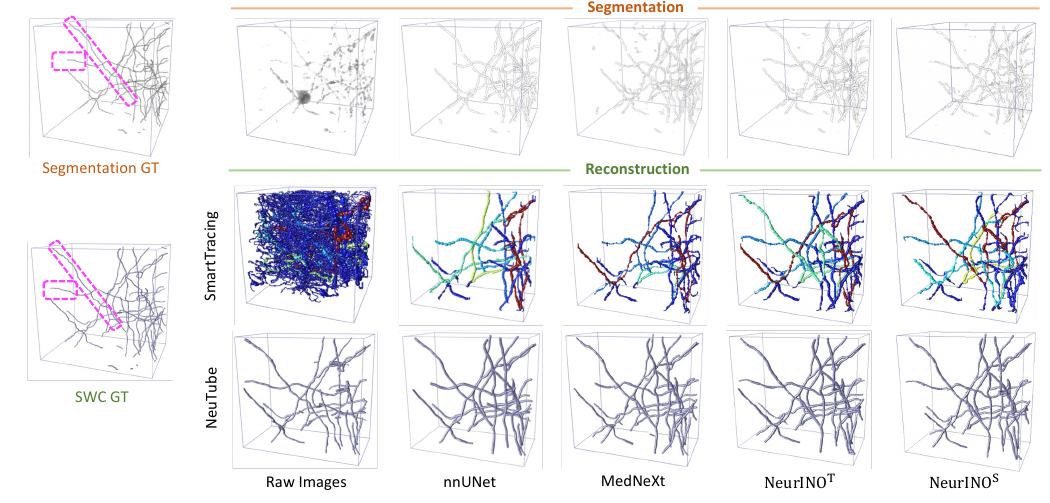}}
    \vspace{-12pt}
   \caption{\textbf{Visualization comparison of the segmentation and tracing results on the CWMBS dataset.} The top row presents raw images and segmentation outputs, followed by two rows showing the corresponding reconstruction results generated by SmartTracing and NeuTube. \textcolor{magenta}{\textbf{Magenta boxes}} indicate severe false negatives (missed neurites) in other methods. Best viewed in zoom-in regions.}
   \label{fig:visual_CWMBS}
   \vspace{-15pt}
\end{figure*}


\section{Experiments and Results} 

\subsection{Datasets}

We evaluate on four publicly available volumetric neuroimaging datasets: two datasets from BigNeuron \cite{peng2015bigneuron} (Drosophila: 42 volumes, 38 train / 4 test images, sizes $88\times89\times93$ – $332\times371\times490$; Mouse: 22 volumes, 19 train / 3 test images, sizes $30\times800\times800$ – $100\times2048\times2048$), NeuroFly \cite{zhao2024neurofly} (constructed from VISoR \cite{wang2019scalable} and fMOST \cite{gong2016high} imaging; 153 volumes, 91 train / 62 test images, each $128\times128\times128$), and CWMBS \cite{liu2024brain} (245 mouse brain volumes: 83 with strong background noise of 49 train / 34 test, and 162 with weak fluorescence of 97 train / 65 test; each $256\times256\times256$). Preprocessing and augmentation details are provided in the Supplementary Material.




\subsection{Evaluation Metrics}

We evaluate performance with both segmentation and reconstruction criteria. For segmentation, we report the F1 score and the 95th percentile Hausdorff Distance (HD95), capturing voxel-wise and boundary accuracy. For neuron reconstruction, we use the distance measurements of Entire Structure Average (ESA), Different Structure Average (DSA), and Percentage of Different Structure (PDS) to assess geometric and topological fidelity between predicted and reference traces. Detailed metric definitions are provided in the Supplementary Material.


\subsection{Implementation Details} 

All experiments are conducted on two NVIDIA RTX 4060 Ti GPUs (16 GB each).
Networks are trained for 110 epochs using the AdamW optimizer with an initial learning rate of 0.001.
Automatic mixed precision (AMP) training is used to accelerate computation and reduce memory overhead. 
Following~\cite{buglakova2025tiling}, we replace all decoder normalization layers with Batch Renormalization~\cite{ioffe2017batch} 
to alleviate tiling artifacts during patch-wise inference.  
During inference, a sliding-window strategy with the same patch size as in training is used to efficiently process large 3D volumes. For model initialization, {NeurINO$^{T}$} and {NeurINO$^{S}$} correspond to the Tiny and Small variants of our framework, initialized from the pretrained {DINOv3 ConvNeXt-Tiny} and {DINOv3 ConvNeXt-Small} backbones, respectively.

\subsection{Results}

\subsubsection{Segmentation Evaluation}
Table~\ref{tab:seg_only_f1_hd95} compares segmentation performance across four public neuroimaging datasets. 
Our models substantially outperform other strong-performing segmentation baselines on F1-scores and HD95 across all test splits. 
Between the two variants, {NeurINO$^{S}$ (initialized from DINOv3-Small)} achieves slightly higher F1-scores and lower HD95 in most cases, 
indicating that the smaller vision backbone provides limited intra-slice visual cues. 
However, the margin between NeurINO$^{S}$ and NeurINO$^{T}$ remains small, suggesting that the inflation-based adaptation strategy effectively transfers semantic priors even to the Tiny encoder. These results confirm that large-scale pretrained representations can be efficiently repurposed for 3D volumetric neuron segmentation, thereby reducing reliance on scarce neuronal data.  

\subsubsection{Reconstruction Evaluation}
We evaluate reconstruction performance using {SmartTracing} and {NeuTube} in Table~\ref{tab:tracing_results}. 
Across all datasets, both NeurINO$^{T}$ and NeurINO$^{S}$ achieve lower tracing errors than other competing baselines, demonstrating improved topological coherence and skeleton continuity. Notably, while NeurINO$^{S}$ yields slightly better voxel-level segmentation results as shown in Table~\ref{tab:seg_only_f1_hd95}, 
{NeurINO$^{T}$ achieves superior reconstruction performance in most tracing benchmarks}, except for the Mouse dataset where NeurINO$^{S}$ attains higher reconstruction accuracy. 
This suggests that a smaller encoder, when combined with inflation-based adaptation and topology-aware skeleton loss, may generalize better to global neuronal morphology without overfitting to local texture cues. In other words, scaling up the encoder improves local segmentation fidelity but does not necessarily improve geometric accuracy of reconstructed skeletons. These observations highlight the distinction between voxel-level accuracy and topological faithfulness in neuron reconstruction, offering insights into structure-aware learning.

\vspace{-15pt}
\paragraph{Qualitative Visualization.}
Figures \ref{fig:visual_Fly} - \ref{fig:visual_CWMBS} present qualitative comparisons of neuron segmentation and reconstruction across different datasets. 
Compared with other segmentation baselines, our method produces {more continuous and faithful reconstructed structures}, 
particularly in thin and weak-signal branches, where competing methods often yield misconnected or inconsistent arbors.
NeurINO$^{T}$ exhibits robust generalization to diverse neuronal morphologies, while NeurINO$^{S}$ generates finer boundary delineation and smoother surface continuity.    
{Magenta boxes} and {arrows}  in the figures mark false negatives (missed neurites) and false positives (spurious branches), respectively. 
As shown in the Mouse dataset, NeurINO suppresses redundant spurious extensions that other methods tend to hallucinate, while in other three datasets, it notably reduces missed fine branches compared with nnUNet and MedNeXt. 
These visual results are consistent with the quantitative findings, demonstrating that integrating large-scale DINOv3 priors 
with topology-aware learning enhances both {local morphological precision} and {global structural consistency}.


\begin{table}[t]
\renewcommand{\arraystretch}{1.3}
\setlength{\tabcolsep}{5pt}
\centering
\caption{
\textbf{Effect of the inflation strategies}. Best results are highlighted in \textcolor{red!70!black}{\textbf{red}}.
}
\label{tab:inflation}
\resizebox{\linewidth}{!}{
\begin{tabular}{lcc|ccc|ccc}
\toprule
\multirow{2}{*}{\textbf{Inflation}} 
& \multirow{2}{*}{\textbf{F1 (\%)}} 
& \multirow{2}{*}{\textbf{HD95}} 
& \multicolumn{3}{c|}{\textit{\textbf{SmartTracing}}} 
& \multicolumn{3}{c}{\textit{\textbf{NeuTube}}} \\
\cmidrule(lr){4-6} \cmidrule(lr){7-9}
& & & ESA & DSA & PDS & ESA & DSA & PDS \\
\midrule
Average 
& 49.64 & 3.15 
& 1.71 & 4.34 & 0.21 & 1.90 & 4.40 & 0.23 \\
{Center} 
& \textcolor{red!70!black}{\textbf{50.06}} & \textcolor{red!70!black}{\textbf{3.07}} 
& \textcolor{red!70!black}{\textbf{1.62}} & \textcolor{red!70!black}{\textbf{4.29}} & \textcolor{red!70!black}{\textbf{0.20}} 
& \textcolor{red!70!black}{\textbf{1.75}} & \textcolor{red!70!black}{\textbf{4.18}} & \textcolor{red!70!black}{\textbf{0.21}} \\
\bottomrule
\end{tabular}}
\end{table}

\begin{table}[t]
\renewcommand{\arraystretch}{1.3}
\setlength{\tabcolsep}{5pt}
\centering
\caption{
\textbf{Effect of topology-aware skeleton loss}.
Best results are highlighted in \textcolor{red!70!black}{\textbf{red}}. 
}
\label{tab:tagl}
\resizebox{\linewidth}{!}{
\begin{tabular}{lcc|ccc|ccc}
\toprule
\multirow{2}{*}{\textbf{TASL}} 
& \multirow{2}{*}{\textbf{F1 (\%)}} 
& \multirow{2}{*}{\textbf{HD95}} 
& \multicolumn{3}{c|}{\textit{\textbf{SmartTracing}}} 
& \multicolumn{3}{c}{\textit{\textbf{NeuTube}}} \\
\cmidrule(lr){4-6}\cmidrule(lr){7-9}
& & & ESA & DSA & PDS & ESA & DSA & PDS  \\
\midrule
{w/o TASL}
& \textcolor{red!70!black}{\textbf{50.59}} & \textcolor{red!70!black}{\textbf{2.99}} 
& 1.68 & 4.41 & 0.21 & 1.94 & 4.48 & 0.23 \\
{with TASL}
& 50.06 & 3.07 
& \textcolor{red!70!black}{\textbf{1.62}} & \textcolor{red!70!black}{\textbf{4.29}} & \textcolor{red!70!black}{\textbf{0.20}} 
& \textcolor{red!70!black}{\textbf{1.75}} & \textcolor{red!70!black}{\textbf{4.18}} & \textcolor{red!70!black}{\textbf{0.21}} \\
\bottomrule
\end{tabular}}
\end{table}
\begin{table}[t]
\renewcommand{\arraystretch}{1.3}
\setlength{\tabcolsep}{5pt}
\centering
\caption{
\textbf{Effect of training strategies}.
Best results are highlighted in \textcolor{red!70!black}{\textbf{red}}, and the second best in \textcolor{blue!70!black}{\textbf{blue}}.
}
\label{tab:training}
\resizebox{\linewidth}{!}{
\begin{tabular}{lcc|ccc|ccc}
\toprule
\multirow{2}{*}{\textbf{Strategy}} 
& \multirow{2}{*}{\textbf{F1 (\%)}} 
& \multirow{2}{*}{\textbf{HD95}} 
& \multicolumn{3}{c|}{\textit{\textbf{SmartTracing}}} 
& \multicolumn{3}{c}{\textit{\textbf{NeuTube}}} \\
\cmidrule(lr){4-6}\cmidrule(lr){7-9}
& & & ESA  & DSA  & PDS  & ESA  & DSA  & PDS  \\
\midrule
{Frozen Encoder}
& {{47.22}} & 5.89 
& 1.78 & 4.49 & \textcolor{blue!70!black}{\textbf{0.23}} 
& 2.04 & 4.59 & \textcolor{blue!70!black}{\textbf{0.24}} \\
{From Scratch}
& \textcolor{blue!70!black}{\textbf{48.56}} & \textcolor{blue!70!black}{\textbf{3.72}} 
& \textcolor{blue!70!black}{\textbf{1.76}} & \textcolor{blue!70!black}{\textbf{4.37}} & \textcolor{blue!70!black}{\textbf{0.23}} 
& \textcolor{blue!70!black}{\textbf{1.93}} & \textcolor{blue!70!black}{\textbf{4.36}} & 0.25 \\
{Fine-Tuning}
& \textcolor{red!70!black}{\textbf{50.06}} & \textcolor{red!70!black}{\textbf{3.07}} 
& \textcolor{red!70!black}{\textbf{1.62}} & \textcolor{red!70!black}{\textbf{4.29}} & \textcolor{red!70!black}{\textbf{0.20}} 
& \textcolor{red!70!black}{\textbf{1.75}} & \textcolor{red!70!black}{\textbf{4.18}} & \textcolor{red!70!black}{\textbf{0.21}} \\
\bottomrule
\end{tabular}}
\vspace{-15pt}
\end{table}

\subsection{Ablation Study}

All ablation experiments are conducted on the {Drosophila} dataset to analyze the contribution of each design component, including the inflation strategy, topology-aware skeleton loss, and training strategy. 
Segmentation is evaluated by F1 and HD95, while reconstruction metrics are reported under {SmartTracing} and {NeuTube} in terms of ESA, DSA, and PDS. 
Results are summarized in Tables~\ref{tab:inflation}--\ref{tab:training}.


\noindent
\textbf{Effects of Inflation Strategies.} 
Table~\ref{tab:inflation} shows that replacing \textit{center inflation} with \textit{average inflation} consistently decreases both segmentation and reconstruction accuracy. 
The center-based scheme embeds each 2D filter at the central slice of the 3D kernel, thereby preserving the spatial anchoring between the 2D receptive field and its 3D extension. 
This geometric alignment maintains the original 2D semantic structure when projected into 3D space, resulting in more coherent representation learning and smoother feature transitions across slices. 
In contrast, average inflation replicates the 2D kernel along the depth dimension and performs a {depth-wise averaging}, producing spatially uniform responses that dilute depth semantics and weaken cross-slice correspondence, ultimately reducing the network’s ability to model coherent volumetric features.



\noindent
\textbf{Effect of Topology-aware Skeleton Learning.} 
Table~\ref{tab:tagl} examines the influence of the proposed TASL. 
Removing TASL slightly improves voxel-level segmentation metrics, but leads to a noticeable degradation in reconstruction accuracy across all metrics. 
This observation suggests that TASL encourages the network to preserve global connectivity and morphological completeness of neuronal arbors, prioritizing structural fidelity over local voxel precision. 
In essence, TASL regularizes segmentation learning toward topology-consistent volumetric representations that are better suited for neuronal tracing.

\noindent
\textbf{Effects of Training Strategies.}
Table~\ref{tab:training} compares different training strategies. 
Freezing the pretrained DINOv3 encoder leads to a substantial performance drop, confirming that full finetuning is crucial for adapting semantic priors to target neuronal distribution. 
Without encoder adaptation, the model fails to align pre-trained visual representations with the fine-grained filamentary structures present in microscopy volumes. 
Training the network from scratch with random initialization further degrades both segmentation and reconstruction performance, highlighting the importance of large-scale visual pretraining for data-efficient 3D neuron reconstruction. 
These results collectively validate that full finetuning of pretrained features enables the network to achieve better structural discrimination and topological consistency across complex neuron morphologies.





\section{Discussion and Conclusion}
We presented \textbf{NeurINO}, a data-efficient framework that adapts the 2D foundation model DINOv3 for 3D neuron segmentation and reconstruction. Through inflation-based adaptation and topology-aware regularization, NeurINO achieves morphologically faithful results with limited annotations.  While this method offers an effective compromise between pretraining scalability and 3D specificity, it remains an approximation to true volumetric understanding. Future research could explore {genuinely 3D foundation models} pretrained on large-scale neuroimaging datasets, capable of learning volumetric context and neural topology from 3D data. Broader open access to large 3D volumes will be crucial, as this would support scalable pretraining and accelerate the emergence of universal 3D neuronal models.

\section*{Acknowledgements}
We thank Xuanhua Yin for helpful feedback and suggestions on improving the manuscript.

{
    \small
    \bibliographystyle{ieeenat_fullname}
    \bibliography{main}
}

\clearpage
\maketitlesupplementary 

\noindent
This supplementary material provides additional technical details and extended results supporting the main paper. 
Specifically, we include: 
(i) definitions of the Dice and Cross-Entropy losses used in training, presented in Section~\ref{sec:loss};
(ii) data preprocessing and augmentation procedures, described in Section~\ref{sec:data_aug};
(iii) formal descriptions of segmentation and reconstruction metrics, detailed in Section~\ref{sec:metrics};
(iv) additional implementation details, including the modification of the DINOv3 downsampling stem, and the hyperparameter settings of TASL, provided in Section~\ref{sec:impl_details};
(v) full quantitative results of additional baselines (TransUNet and SegMamba), reported in Section~\ref{sec:suppl_results};
(vi) cross-method evaluation of TASL on additional segmentation backbones in Section~\ref{sec:cross_method};
(vii) computational analysis, indicated in Section~\ref{sec:comp};
(viii) an analysis of pretraining effects on training effectiveness and feature behavior in Section~\ref{sec:pretrain};
(ix) component-wise ablation of TASL in Section~\ref{sec:tasl_ablation}; and
(x) additional qualitative visualizations across four neuron datasets, shown in Section~\ref{sec:visuals}.




\section{Additional Loss Definitions}
\label{sec:loss}
Following the main paper, our total objective combines Dice and Cross-Entropy (CE) losses. 
Below we provide the definitions of the Dice and CE losses.  

\noindent\textbf{Dice Loss.}
The Dice loss is a function that measures the overlap between the prediction and the ground truth. It is defined as:

\begin{equation}
\mathcal{L}_{\text{Dice}} = 1 - \frac{2 \sum_{i} p_i g_i}{\sum_{i} p_i^2 + \sum_{i} g_i^2 + \epsilon},
\end{equation}

\noindent where $p_i$ and $g_i$ represent the predicted probability and ground-truth label at voxel $i$, and $\epsilon$ is a small constant used to prevent division by zero.

\noindent\textbf{Cross-Entropy Loss.}
The Cross-Entropy loss is employed to ensure voxel-wise classification accuracy:

\begin{equation}
\mathcal{L}_{\text{CE}} = - \sum_{i} g_i \log p_i + (1 - g_i) \log (1 - p_i).
\end{equation}

\section{Data Preprocessing and Augmentation}
\label{sec:data_aug}
To ensure robust training and good generalization, we apply the following data preprocessing and augmentation steps:

\begin{itemize}
\item \textbf{Intensity Normalization:} Voxel intensities are normalized to zero mean and unit variance to reduce variability. 
\item \textbf{Patch-Based Cropping:} Due to GPU memory constraints, we adopt a patch-based strategy by randomly extracting sub-volumes such as $32\times32\times32$. 
\item \textbf{Data Augmentation:} To mitigate overfitting, we apply augmentations including random rotation, flipping, scaling, cropping, elastic deformation, gamma correction, brightness/contrast jitter, and Gaussian noise injection.
\end{itemize}

\section{Evaluation Metric Definitions}
\label{sec:metrics}

\noindent\textbf{Segmentation Metrics.}
Voxel-wise segmentation performance is evaluated using the F1 score and the 95th percentile Hausdorff Distance (HD95). 
A higher F1 score indicates better foreground prediction, while a lower HD95 reflects smaller boundary deviation between the prediction and the ground truth.

The F1 score is defined as:

\begin{equation}
\text{F1 Score} = \frac{2 \cdot \text{Precision} \cdot \text{Recall}}{\text{Precision} + \text{Recall}},
\end{equation}

\noindent where the Precision measures the proportion of correctly predicted foreground voxels among all voxels predicted as foreground, and Recall quantifies the proportion of actual foreground voxels that are correctly predicted.

The definition of HD95 is presented as follows:
\begin{equation}
\text{HD95} =
\operatorname*{P_{95}}\Big(
\min_{y \in \mathcal{V}_{\text{gt}}} \|x - y\|
\Big),
\quad x \in \mathcal{V}_{\text{pred}},
\end{equation}
where 
$\mathcal{V}_{\text{pred}}$ and $\mathcal{V}_{\text{gt}}$ denote the surface voxel sets of the prediction and the ground truth, respectively,  
$\|\cdot\|$ is the Euclidean distance,  
and $P_{95}(\cdot)$ extracts the 95th percentile.


\begin{table*}[t]
\centering
\scriptsize
\setlength{\tabcolsep}{10pt}
\renewcommand{\arraystretch}{1}
\caption{
\textbf{Quantitative comparison of neuron segmentation performance.}
Each cell reports {F1-score (\%) / HD95} (lower is better for HD95) across four datasets.
Columns correspond to datasets and rows to different segmentation methods.
{NeurINO$^{T}$} and {NeurINO$^{S}$} denote the Tiny and Small variants of NeurINO.
Best results are highlighted in \textcolor{red!70!black}{\textbf{red}}, and the second best in \textcolor{blue!70!black}{\textbf{blue}}.
}
\label{tab:suppl_seg}

\setlength{\heavyrulewidth}{0.06em}
\setlength{\lightrulewidth}{0.05em}
\begin{tabular}{lcccccc}
\toprule
\textbf{Method} & \textbf{Source} & \textbf{Params (M)} & \textbf{Drosophila} & \textbf{Mouse} & \textbf{NeuroFly} & \textbf{CWMBS} \\
\midrule
nnUNet~\cite{isensee2021nnu}      & \textit{Nat. Methods 2021} & 27.66 & 47.20 / 3.20 & 52.05 / 10.12 & 63.36 / 18.33 & {{36.50 / 16.34}} \\
MedNeXt~\cite{roy2023mednext}     & \textit{MICCAI 2023} & 61.97 & {47.74 / 3.15} & 50.61 / 13.77 & 62.50 / 19.23 & 33.46 / 18.37 \\
TransUNet~\cite{chen2024transunet}& \textit{MIA 2024} & 105.28 & 34.88 / 16.67 & 38.43 / 48.13 & 37.93 / 18.55 & 24.56 / 21.14 \\
SegMamba~\cite{xing2024segmamba}  & \textit{MICCAI 2024} & 67.42 & 37.48 / 5.03 & 39.98 / 56.37 & 46.77 / 17.12 & 28.01 / 19.57 \\

\midrule
NeurINO$^{T}$ & - & 39.21 & \textcolor{blue!70!black}{\textbf{50.06 / 3.07}} & \textcolor{blue!70!black}{\textbf{52.50 / 9.50}} & \textcolor{blue!70!black}{\textbf{65.23}} / \textcolor{red!70!black}{\textbf{16.38}} & \textcolor{red!70!black}{\textbf{36.77 / 16.10}} \\

NeurINO$^{S}$ & - & 61.52 & \textcolor{red!70!black}{\textbf{50.19 / 3.02}} & \textcolor{red!70!black}{\textbf{52.73 / 9.24}} & \textcolor{red!70!black}{\textbf{65.44}} / \textcolor{blue!70!black}{\textbf{16.53}} & \textcolor{blue!70!black}{\textbf{36.55 / 16.27}} \\
\bottomrule
\end{tabular}
\end{table*}


\begin{table*}[t]
\centering
\small
\setlength{\tabcolsep}{4.5pt}
\renewcommand{\arraystretch}{2.5}
\caption{
\textbf{Quantitative comparison of neuron tracing performance.}
Each cell reports {ESA / DSA / PDS} (lower is better for all metrics) for two tracing algorithms: 
SmartTracing and NeuTube, evaluated across four datasets. 
{NeurINO$^{T}$} and {NeurINO$^{S}$} denote the Tiny and Small variants of NeurINO.
Best results are highlighted in \textcolor{red!70!black}{\textbf{red}}, and the second best in \textcolor{blue!70!black}{\textbf{blue}}. 
}
\label{tab:suppl_tracing}
\resizebox{\textwidth}{!}{
\fontsize{15}{10}\selectfont
\begin{tabular}{lcccccccc}
\toprule
\multirow{2}{*}{\textbf{Method}} &
\multicolumn{2}{c}{\textbf{Drosophila}} &
\multicolumn{2}{c}{\textbf{Mouse}} &
\multicolumn{2}{c}{\textbf{NeuroFly}} &
\multicolumn{2}{c}{\textbf{CWMBS}} \\
\cmidrule(lr){2-3} \cmidrule(lr){4-5} \cmidrule(lr){6-7} \cmidrule(lr){8-9}
& \textit{SmartTracing} & \textit{NeuTube}
& \textit{SmartTracing} & \textit{NeuTube}
& \textit{SmartTracing} & \textit{NeuTube}
& \textit{SmartTracing} & \textit{NeuTube} \\
\midrule

nnUNet~\cite{isensee2021nnu}
& 1.67 / 4.48 / \textcolor{red!70!black}{\textbf{0.20}} 
& 1.87 / 4.67 / 0.24
& 2.99 / 8.06 / \textcolor{blue!70!black}{\textbf{0.22}} 
& 5.36 / 16.10 / \textcolor{blue!70!black}{\textbf{0.22}}
& \textcolor{blue!70!black}{\textbf{22.78}} / 31.76 / 0.41 
& \textcolor{blue!70!black}{\textbf{4.22}} / 14.13 / \textcolor{red!70!black}{\textbf{0.15}}
& 36.93 / 42.10 / \textcolor{red!70!black}{\textbf{0.53}} 
& 8.02 / 14.73 / \textcolor{blue!70!black}{\textbf{0.41}} \\

MedNeXt~\cite{roy2023mednext}
& 1.68 / 4.44 / \textcolor{red!70!black}{\textbf{0.20}} 
& 1.90 / 4.61 / 0.23
& 4.06 / 11.66 / 0.25 
& 6.81 / 19.61 / \textcolor{blue!70!black}{\textbf{0.22}}
& 27.52 / 34.74 / 0.40 
& 4.32 / 14.06 / 0.17
& 34.74 / 38.94 / 0.56 
& 7.93 / \textcolor{blue!70!black}{\textbf{13.58}} / 0.44 \\

TransUNet~\cite{chen2024transunet}
& 20.88 / 22.97 / 0.48 
& 4.59 / 9.11 / 0.38
& 25.99 / 33.25 / 0.64 
& 22.62 / 32.92 / 0.61
& 27.26 / 34.84 / 0.57 
& 11.11 / 17.24 / 0.39
& 38.40 / 43.14 / 0.61 
& 13.62 / 20.55 / 0.44 \\

SegMamba~\cite{xing2024segmamba}
& 43.58 / 46.37 / 0.46 
& 5.14 / 8.23 / 0.32
& 30.28 / 41.19 / 0.55 
& 35.41 / 49.75 / 0.55
& 22.82 / 31.53 / 0.47 
& 8.97 / 18.53 / 0.29
& 37.54 / 42.56 / 0.57 
& 10.57 / 17.70 / \textcolor{red!70!black}{\textbf{0.40}} \\

\midrule
NeurINO$^{T}$
& \textcolor{red!70!black}{\textbf{1.62}} / \textcolor{red!70!black}{\textbf{4.29}} / \textcolor{red!70!black}{\textbf{0.20}} 
& \textcolor{red!70!black}{\textbf{1.75}} / \textcolor{red!70!black}{\textbf{4.18}} / \textcolor{red!70!black}{\textbf{0.21}}
& \textcolor{red!70!black}{\textbf{2.81}} / \textcolor{blue!70!black}{\textbf{7.88}} / \textcolor{blue!70!black}{\textbf{0.22}} 
& \textcolor{blue!70!black}{\textbf{4.86}} / \textcolor{red!70!black}{\textbf{15.30}} / \textcolor{blue!70!black}{\textbf{0.22}}
& \textcolor{red!70!black}{\textbf{21.93}} / \textcolor{red!70!black}{\textbf{28.98}} / \textcolor{red!70!black}{\textbf{0.32}} 
& \textcolor{red!70!black}{\textbf{4.13}} / \textcolor{red!70!black}{\textbf{13.91}} / \textcolor{blue!70!black}{\textbf{0.16}}
& \textcolor{blue!70!black}{\textbf{34.09}} / \textcolor{red!70!black}{\textbf{37.76}} / \textcolor{blue!70!black}{\textbf{0.54}} 
& \textcolor{red!70!black}{\textbf{7.74}} / \textcolor{red!70!black}{\textbf{13.40}} / \textcolor{blue!70!black}{\textbf{0.41}} \\

NeurINO$^{S}$
& \textcolor{blue!70!black}{\textbf{1.65}} / \textcolor{blue!70!black}{\textbf{4.40}} / \textcolor{blue!70!black}{\textbf{0.21}} 
& \textcolor{blue!70!black}{\textbf{1.84}} / \textcolor{blue!70!black}{\textbf{4.31}} / \textcolor{blue!70!black}{\textbf{0.22}}
& \textcolor{blue!70!black}{\textbf{2.92}} / \textcolor{red!70!black}{\textbf{7.73}} / \textcolor{red!70!black}{\textbf{0.21}} 
& \textcolor{red!70!black}{\textbf{4.73}} / \textcolor{blue!70!black}{\textbf{15.53}} / \textcolor{red!70!black}{\textbf{0.21}}
& 23.61 / \textcolor{blue!70!black}{\textbf{30.84}} / \textcolor{blue!70!black}{\textbf{0.36}} 
& 4.24 / \textcolor{blue!70!black}{\textbf{14.02}} / \textcolor{red!70!black}{\textbf{0.15}}
& \textcolor{red!70!black}{\textbf{33.48}} / \textcolor{blue!70!black}{\textbf{38.09}} / \textcolor{blue!70!black}{\textbf{0.54}} 
& \textcolor{blue!70!black}{\textbf{7.89}} / 13.67 / \textcolor{blue!70!black}{\textbf{0.41}} \\
\bottomrule
\end{tabular}
}
\vspace{-15pt}
\end{table*}


\noindent\textbf{Reconstruction Metrics.}
We adopt three standard tracing metrics widely used in neuron reconstruction:  
Entire Structure Average (ESA), Different Structure Average (DSA), and Percentage of Different Structure (PDS). 
Lower values of all three metrics indicate better topological consistency and reconstruction accuracy. 

Given the predicted skeleton node set $\mathcal{S}_{\text{pred}}$ and the ground-truth node set $\mathcal{S}_{\text{gt}}$,  
ESA measures the average minimum distance from each node in the prediction to its closest node in the ground truth:
\begin{equation}
\text{ESA} = 
\frac{1}{|\mathcal{S}_{\text{pred}}|}
\sum_{x \in \mathcal{S}_{\text{pred}}}
\min_{y \in \mathcal{S}_{\text{gt}}} \| x - y \|.
\end{equation} 

DSA measures the mean distance between non-overlapping regions:
\begin{equation}
\text{DSA} = 
\frac{1}{|\mathcal{S}_{\text{pred}}'|}
\sum_{x \in \mathcal{S}_{\text{pred}}'} 
\min_{y \in \mathcal{S}_{\text{gt}}'} \| x - y \|,
\end{equation}
where $\mathcal{S}_{\text{pred}}'$ and $\mathcal{S}_{\text{gt}}'$ represent the non-overlapping skeleton nodes of the prediction and the ground truth.

PDS measures the proportion of mismatched branches over the total branch length:
\begin{equation}
\text{PDS} = 
\frac{\text{Length of mismatched branches}}
{\text{Total branch length}}.
\end{equation}


\vspace{-3pt}
\section{Additional Implementation Details}
\label{sec:impl_details} 

\textbf{Modification of the DINOv3 Downsampling Stem.}
The original DINOv3 ConvNeXt-style downsampling stem performs a $4\times$ spatial reduction (stride~4) in the first layer. 
While suitable for 2D natural images, such aggressive resolution reduction tends to eliminate fine neuronal structures that are only a few voxels wide. 
To better preserve high-frequency structural cues critical for neurite continuity, we replace the stride-4 stem with a stride-2 variant and apply appropriate padding to maintain spatial alignment during 3D convolution, resulting in a $2\times$ downsampling at the input stage. 
This modification substantially improves the retention of thin neurites during early feature extraction and subsequently enhances both segmentation and reconstruction performance, as reported in Table~\ref{tab:downsample}.

\noindent\textbf{Hyperparameters of TASL.}
For TASL, we use the weighting coefficients $\lambda_{\text{node}}$, $\lambda_{\text{edge}}$, and $\lambda_{\text{path}}$ for the node-, edge-, and path-level terms, respectively.
These coefficients are set to 1.0, 0.5, and 0.5. They are kept fixed in all experiments.

\begin{table}  
\renewcommand{\arraystretch}{1.3}
\setlength{\tabcolsep}{5pt}
\centering
\caption{
\textbf{Effect of modifying the first downsampling ratio in DINOv3.} 
Better results are highlighted in \textcolor{red!70!black}{\textbf{red}}.
}
\label{tab:downsample}
\resizebox{\linewidth}{!}{
\begin{tabular}{lcc|ccc|ccc}
\toprule
\multirow{2}{*}{\shortstack[l]{\textbf{Downsampling} \\ \textbf{Ratio}}}
& \multirow{2}{*}{\textbf{F1 (\%)}} 
& \multirow{2}{*}{\textbf{HD95}} 
& \multicolumn{3}{c|}{\textit{\textbf{SmartTracing}}} 
& \multicolumn{3}{c}{\textit{\textbf{NeuTube}}} \\
\cmidrule(lr){4-6} \cmidrule(lr){7-9}
& & & ESA & DSA & PDS & ESA & DSA & PDS \\
\midrule
4$\times$
& 47.08 & 4.86
& 1.80 & \textcolor{red!70!black}{\textbf{4.25}} & 0.24 
& 1.98 & 4.47 & 0.25 \\

2$\times$
& \textcolor{red!70!black}{\textbf{50.06}} 
& \textcolor{red!70!black}{\textbf{3.07}}
& \textcolor{red!70!black}{\textbf{1.62}} 
& 4.29 
& \textcolor{red!70!black}{\textbf{0.20}}
& \textcolor{red!70!black}{\textbf{1.75}} 
& \textcolor{red!70!black}{\textbf{4.18}} 
& \textcolor{red!70!black}{\textbf{0.21}} \\
\bottomrule
\end{tabular}}
\vspace{-6pt}
\end{table}


\begin{table} 
\renewcommand{\arraystretch}{1.3}
\setlength{\tabcolsep}{5pt}
\centering
\caption{
\textbf{Cross-method evaluation of TASL.}
Best results are highlighted in \textbf{bold}. 
}
\label{tab:cross_method}
\resizebox{\linewidth}{!}{
\begin{tabular}{lcc|ccc|ccc}
\toprule
\multirow{2}{*}{\textbf{Method}} 
& \multirow{2}{*}{\textbf{F1 (\%)}} 
& \multirow{2}{*}{\textbf{HD95}} 
& \multicolumn{3}{c|}{\textit{\textbf{SmartTracing}}} 
& \multicolumn{3}{c}{\textit{\textbf{NeuTube}}} \\
\cmidrule(lr){4-6}\cmidrule(lr){7-9}
& & & ESA  & DSA  & PDS  & ESA  & DSA  & PDS  \\
\midrule
nnUNet 
& \textbf{47.20} & 3.20
& 1.67 & 4.48 & \textbf{0.20}
& 1.87 & 4.67 & 0.24 \\
nnUNet + TASL 
& 47.09  &  \textbf{3.14}
& \textbf{1.65} & \textbf{4.41} & 0.21 
& \textbf{1.84} & \textbf{4.52} & \textbf{0.23} \\
\midrule
MedNeXt  
& \textbf{47.74} & 3.15
& 1.68 & 4.44 & \textbf{0.20}
& 1.90 & 4.61 & \textbf{0.23} \\
MedNeXt + TASL 
& 47.66 &  \textbf{3.11}
& \textbf{1.65} & \textbf{4.36} &  0.21
& \textbf{1.85} & \textbf{4.40} & 0.24 \\
\bottomrule
\end{tabular}}
\vspace{-10pt}
\end{table}


\section{Full Quantitative Results}
\label{sec:suppl_results} 

Due to space limitations, the main paper reports a subset of baseline models.
Here we provide the complete quantitative results, including additional baselines TransUNet~\cite{chen2024transunet} and SegMamba~\cite{xing2024segmamba}, as illustrated in Table~\ref{tab:suppl_seg} and Table~\ref{tab:suppl_tracing}.

\vspace{-5pt}
\section{Cross-method Evaluation of TASL}
\label{sec:cross_method}
To evaluate the generality of TASL beyond the proposed NeurINO architecture,
we integrate TASL into two representative 3D segmentation backbones,
nnUNet and MedNeXt.
Table~\ref{tab:cross_method} reports the results.
TASL consistently improves topology-sensitive reconstruction metrics,
demonstrating that its benefits are not limited to our architecture.

\begin{figure}
    \centering
    \includegraphics[width=0.95\linewidth]
    {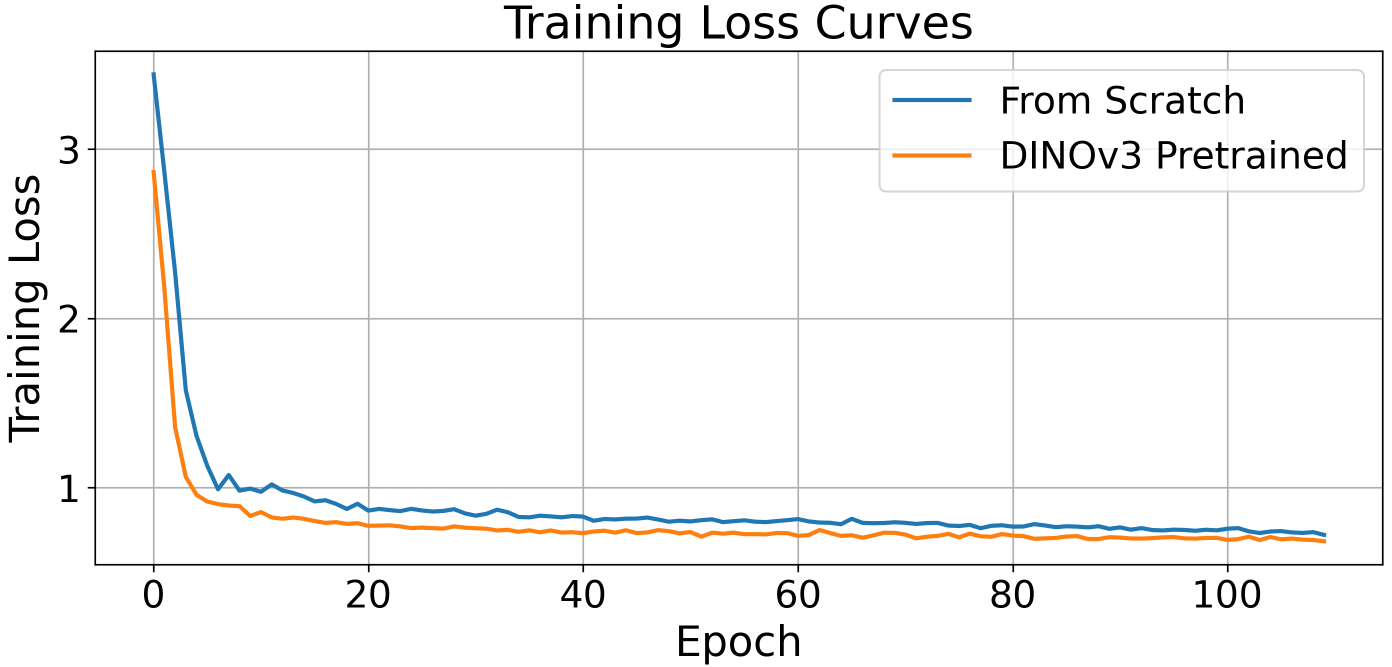}
    \caption{
    \textbf{Training loss curves comparing models trained from scratch and initialized with DINOv3 pretrained weights.}
    Pretraining accelerates convergence and stabilizes training dynamics.
    }
    \label{fig:training_curve}
\vspace{-6pt}  
\end{figure}

\begin{figure}
    \centering
    \includegraphics[width=0.95\linewidth]
    {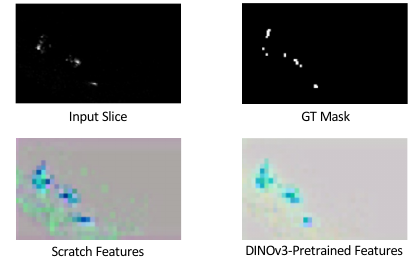}
    \vspace{-4pt}
    \caption{
    \textbf{ Visualization of encoder features with and without DINOv3 pretraining.}
    Compared with features learned from scratch, DINOv3-pretrained features exhibit more coherent responses and clearer separation from the background.
    }
    \label{fig:feature_vis}
\vspace{-10pt}  
\end{figure}


\begin{figure*}
  \centering
  \centerline{\includegraphics[width=16.5cm]{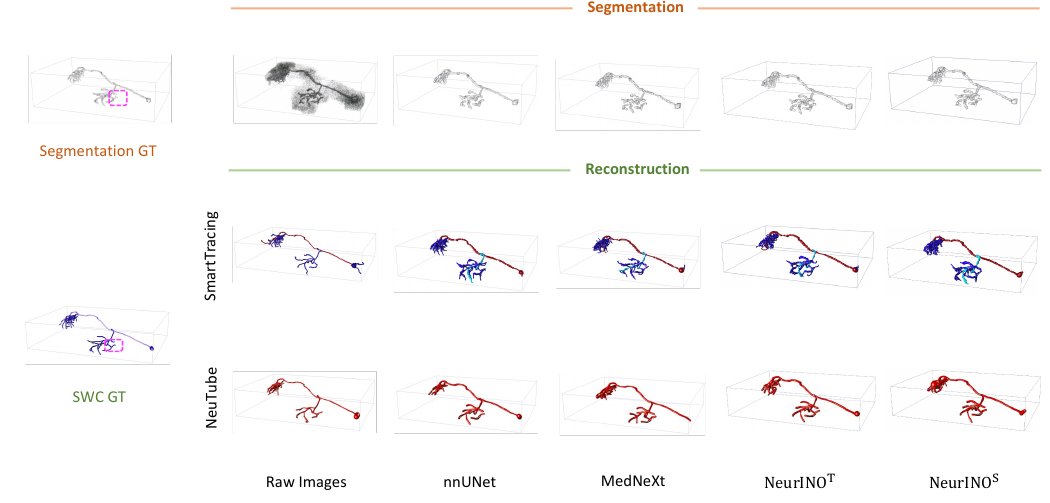}}
    \vspace{-6pt}
   \caption{\textbf{Visualization comparison of the segmentation and tracing results on the Drosophila dataset.} The top row presents raw images and segmentation outputs, followed by two rows showing the corresponding reconstruction results generated by SmartTracing and NeuTube. \textcolor{magenta}{\textbf{Magenta boxes}} indicate severe false negatives (missed neurites) in other methods. Best viewed in zoom-in regions.} 
   \label{fig:Suppl_visual1}
   \vspace{-4pt}
\end{figure*}

\begin{figure*}
  \centering
  \centerline{\includegraphics[width=17cm]{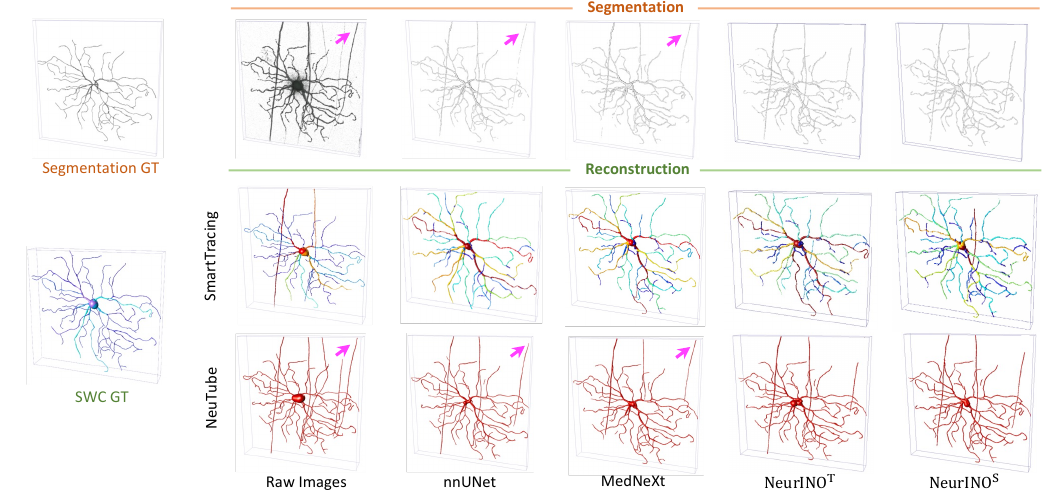}}
    \vspace{-6pt}
   \caption{\textbf{Visualization comparison of the segmentation and tracing results on the Mouse dataset.} The top row presents raw images and segmentation outputs, followed by two rows showing the corresponding reconstruction results generated by SmartTracing and NeuTube. \textcolor{Magenta}{\textbf{Magenta arrows}} highlight severe {false positives} in other methods. Best viewed in zoom-in regions.}
   \label{fig:Suppl_visual2}
   \vspace{-15pt}
\end{figure*}

\begin{figure*}
  \centering
  \centerline{\includegraphics[width=17cm]{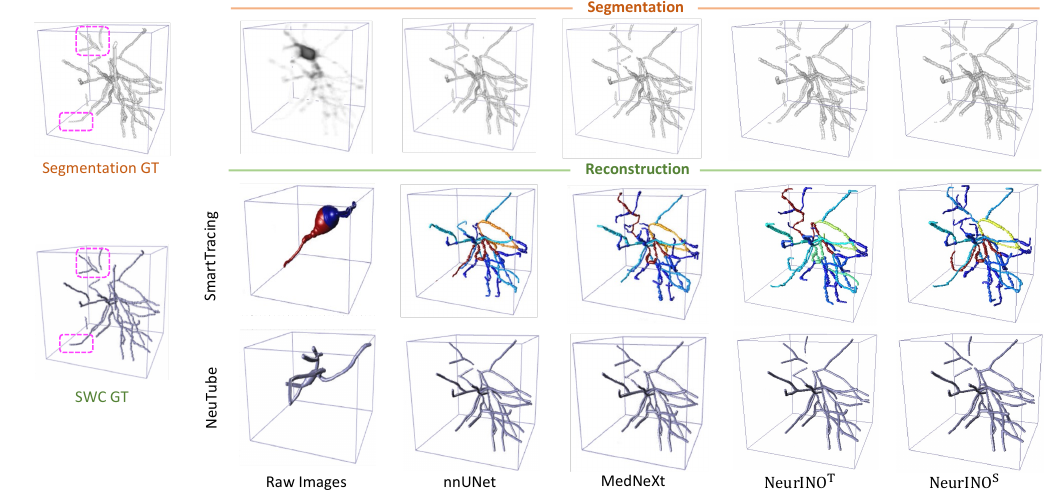}}
    \vspace{-6pt}
   \caption{\textbf{Visualization comparison of the segmentation and tracing results on the NeuroFly dataset.} The top row presents raw images and segmentation outputs, followed by two rows showing the corresponding reconstruction results generated by SmartTracing and NeuTube. \textcolor{magenta}{\textbf{Magenta boxes}} indicate severe false negatives (missed neurites) in other methods. Best viewed in zoom-in regions.} 
   \label{fig:Suppl_visual3}
   \vspace{-4pt}
\end{figure*}

\begin{figure*}
  \centering
  \centerline{\includegraphics[width=17cm]{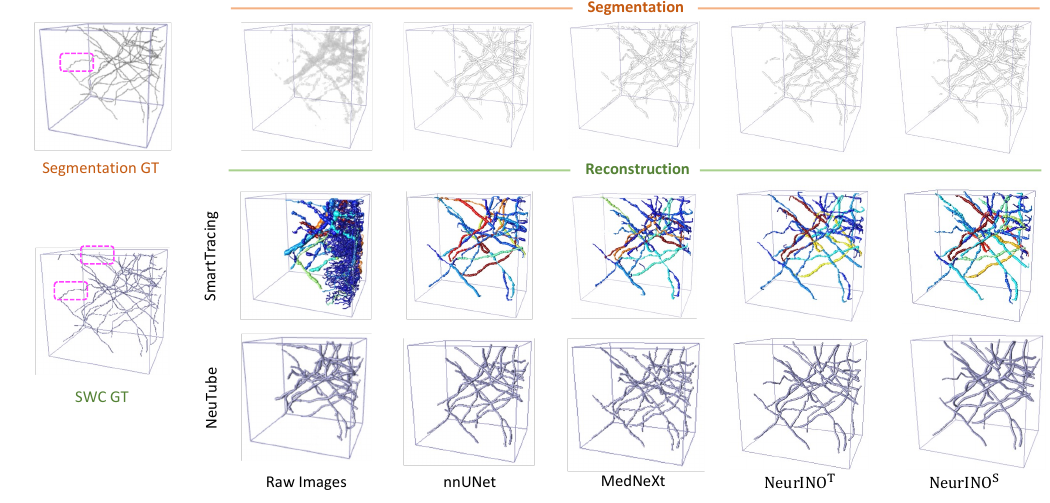}}
    \vspace{-12pt}
   \caption{\textbf{Visualization comparison of the segmentation and tracing results on the CWMBS dataset.} The top row presents raw images and segmentation outputs, followed by two rows showing the corresponding reconstruction results generated by SmartTracing and NeuTube. \textcolor{magenta}{\textbf{Magenta boxes}} indicate severe false negatives (missed neurites) in other methods. Best viewed in zoom-in regions.}
   \label{fig:Suppl_visual4}
   \vspace{-15pt}
\end{figure*}






\vspace{-6pt} 
\section{Effect of Pretraining on Training Effectiveness}
\label{sec:pretrain}
Figure~\ref{fig:training_curve} compares training curves when initializing with DINOv3 pretrained weights versus training the model entirely from scratch.
It is demonstrated that pretrained initialization leads to faster convergence, lower training loss, and smoother optimization, particularly in early training epochs. 

To further analyze the effect of DINOv3 pretraining, we visualize encoder feature representations learned with and without pretraining. 
Given a 3D volume, we select a representative slice containing neuronal structures and extract encoder features. 
The feature maps are projected to three dimensions using principal component analysis (PCA) and visualized as RGB images. 
As shown in Figure~\ref{fig:feature_vis}, features learned from scratch tend to exhibit weaker structural coherence and noisier responses. 
In contrast, DINOv3-pretrained features produce more consistent activations along neuronal branches and better separation from the background. 
These results suggest that pretrained representations provide stronger intra-slice semantic cues, facilitating more effective feature aggregation across slices during training.

\section{Component-wise Ablation of TASL}
\label{sec:tasl_ablation}
To further evaluate the contribution of each term in TASL, 
we conduct a component-wise ablation by removing different combinations of the node-, edge-, and path-level terms.
Table~\ref{tab:tasl_ablation} reports the results.
Using all three components together yields the most balanced performance across segmentation and reconstruction metrics, confirming their complementary roles in preserving neuronal morphology.

\vspace{-6pt}
\begin{table}[H]
\renewcommand{\arraystretch}{1.3}
\setlength{\tabcolsep}{5pt}
\centering
\caption{
\textbf{Ablation of TASL components.}
Best results are highlighted in \textcolor{red!70!black}{\textbf{red}}, and the second best in \textcolor{blue!70!black}{\textbf{blue}}.
}
\label{tab:tasl_ablation}
\resizebox{\linewidth}{!}{
\begin{tabular}{lcc|ccc|ccc}
\toprule
\multirow{2}{*}{\textbf{Variant}} 
& \multirow{2}{*}{\textbf{F1 (\%)}} 
& \multirow{2}{*}{\textbf{HD95}} 
& \multicolumn{3}{c|}{\textit{\textbf{SmartTracing}}} 
& \multicolumn{3}{c}{\textit{\textbf{NeuTube}}} \\
\cmidrule(lr){4-6}\cmidrule(lr){7-9}
& & & ESA  & DSA  & PDS  & ESA  & DSA  & PDS  \\
\midrule
TASL (Node + Edge)
& \textcolor{blue!70!black}{\textbf{49.94}} & 3.22 
& 1.67 & 4.36 & 0.22 
& 1.86 & 4.24 & \textcolor{red!70!black}{\textbf{0.21}} \\
TASL (Node + Path)
& 49.86 & \textcolor{blue!70!black}{\textbf{3.13}} 
& \textcolor{blue!70!black}{\textbf{1.64}} & \textcolor{red!70!black}{\textbf{4.26}} & 0.22
& 1.83 & \textcolor{blue!70!black}{\textbf{4.21}} & 0.22 \\
TASL (Edge + Path)
& 49.75 & 3.20 
& 1.66 & 4.34 & \textcolor{blue!70!black}{\textbf{0.21}}
& \textcolor{blue!70!black}{\textbf{1.82}} & 4.29 & 0.22 \\
TASL (Full)
& \textcolor{red!70!black}{\textbf{50.06}} & \textcolor{red!70!black}{\textbf{3.07}}
& \textcolor{red!70!black}{\textbf{1.62}} & \textcolor{blue!70!black}{\textbf{4.29}} & \textcolor{red!70!black}{\textbf{0.20}}
& \textcolor{red!70!black}{\textbf{1.75}} & \textcolor{red!70!black}{\textbf{4.18}} & \textcolor{red!70!black}{\textbf{0.21}} \\
\bottomrule
\end{tabular}}
\vspace{-4pt}
\end{table}


\section{Computational Analysis}
\label{sec:comp}
We analyze the computational complexity of NeurINO and compare it with representative 3D segmentation baselines.
Table~\ref{tab:comp} reports the number of parameters and floating-point operations (FLOPs) for each model.
All FLOPs are measured for a single forward pass with input size $64\times64\times64$.
Compared with conventional convolutional architectures such as nnUNet and MedNeXt, NeurINO achieves competitive or lower computational cost.
In particular, NeurINO$^{T}$ requires significantly fewer FLOPs while maintaining strong reconstruction performance, demonstrating its efficiency.
Even the larger NeurINO$^{S}$ variant has comparable computational complexity to MedNeXt while providing improved reconstruction quality.


\vspace{-6pt}
\begin{table}[H]
\renewcommand{\arraystretch}{1.3}
\setlength{\tabcolsep}{5pt}
\centering
\footnotesize
\caption{
\textbf{Computational analysis of different models.}
We report the number of parameters and FLOPs for representative segmentation methods.
}
\label{tab:comp}
\begin{tabular}{l@{\hspace{40pt}}c@{\hspace{40pt}}c}
\toprule
\textbf{Model} & \textbf{Params (M)} & \textbf{FLOPs (G)} \\
\midrule
nnUNet & 27.66 & 55.68 \\
MedNeXt & 61.97 & 61.55 \\
NeurINO$^{T}$ & 39.21 & 43.56 \\
NeurINO$^{S}$ & 61.52 & 54.97 \\
\bottomrule
\end{tabular}
\end{table}

\section{Additional Qualitative Visualizations}
\label{sec:visuals}
We provide additional visualization results across four datasets in Figures~\ref{fig:Suppl_visual1}-\ref{fig:Suppl_visual4}. 
Our method yields morphologically continuous reconstructions with fewer broken branches and more faithful neuronal arbor topology.


\end{document}